\documentclass[11pt]{article}
\usepackage{acl2015}
\usepackage{times}
\usepackage{url}
\usepackage{latexsym}
\usepackage{url}
\usepackage{graphicx,comment}
\usepackage{multirow}
\usepackage{arydshln}	
\usepackage[utf8]{inputenc}

\title{Incremental Adaptation Strategies \\ for Neural Network Language Models}

\author{Aram Ter-Sarkisov, Holger Schwenk, Lo\"{\i}c Barrault  and Fethi Bougares \\
  School of Computer Science, University of Maine, \\
  Le Mans, France\\
  {\tt tersarkisov@lium.univ-lemans.fr} \\
  {\tt holger.schwenk@lium.univ-lemans.fr}\\
  {\tt loic.barrault@lium.univ-lemans.fr}\\
  {\tt fethi.bougares@lium.univ-lemans.fr}\\
} 

\date{}

\newcommand{\MC}{\multicolumn}
\newcommand{\MR}{\multirow}
\newcommand{\ra}{\rightarrow}
\newcommand{\eg}{\textit{e.g.~}}
\newcommand{\ie}{\textit{i.e.~}}
\newcommand{\vs}{\mathbf{s}}

\newcommand{\vt}{\mathbf{t}}

\begin{document}
\maketitle

\begin{abstract}
It is today acknowledged that neural network language models outperform
back-off language models in applications like speech recognition or statistical
machine translation.  However, training these models on large amounts of data
can take several days.  We present efficient techniques to adapt a neural
network language model to new data. Instead of training a completely new model
or relying on mixture approaches, we propose two new methods: continued training
on resampled data or insertion of adaptation layers.  We present experimental
results in an CAT environment where the post-edits of professional translators
are used to improve an SMT system.  Both methods are very fast and achieve
significant improvements without over-fitting the small adaptation data.
\end{abstract}


\section{Introduction}

A language model (LM) plays an important role in many natural language
processing applications, namely speech recognition
and statistical machine translation (SMT). For a very long time, back-off
$n$-gram models were considered to be the state-of-the-art, in particular when
large amounts of training data are available.

An alternative approach is based on the use of high-dimensional embeddings of
the words and the idea to perform the probability estimation in this space. By
these means, meaningful interpolations can be expected.  The projection and
probability estimation can be jointly learned by a neural network
\cite{Yoshua:2003:ML}. These models, also called continuous space language
models (CSLM), have seen a surge in popularity, and it was confirmed in many
studies that they systematically outperform back-off $n$-gram models by a
significant margin in SMT and speech recognition.
Many variants of the basic approach were proposed during the last years, e.g.
the use of recurrent architectures~\cite{Mikolov:rcslm:intersp10} or LSTM
\cite{Sundermeyer:2012:is_lstm}.  More recently, neural networks were also
used for the translation model in an SMT system
\cite{Le:2012:naacl,Schwenk:2012:coling,Cho+lium:2014:nnmt_encdec}, and first
translations systems entirely based on neural networks were proposed
\cite{Sutskever:2014:nips_nntrans,bahdanau:14:nipswshop_nnmt}.

However, to the best of our knowledge, all these systems are static, i.e. they
are trained once on a large representative corpus and are not changed or
adapted to new data or conditions.
The ability to adapt to changing conditions is a very important property of an
operational SMT system.  The need for adaptation occurs for instance in a
system to translate daily news articles in order to account for the changing
environment. Another typical application is the integration of an SMT system in
an CAT\footnote{Computer Assisted Translation} tool: we want to improve the SMT
systems with help of user corrections.  Finally, one may also want to adapt a
generic SMT to a particular genre or topic for which we lack large amounts of
specific data.  Various adaptation schemes were proposed for \textit{classical
SMT systems}, but to the best of our knowledge, there is only very limited
works involving neural network models.

We are interested in a setting where an LM needs to be adapted to a small
amount of data which is representative of a domain change, so that the overall
system will perform better on this domain in the future.  Our task, which
corresponds to concrete needs in real-world applications, is the translation of
a document by an human over several days. The human translator is assisted by
an SMT system which proposes translation hypothesis to speed up his work (post
editing). After one day of work, we adapt the CSLM to the translations already
performed by the human translator, and show that the SMT system performs better
on the remaining part of the document.

In this paper, we use the open-source MateCat
tool\footnote{https://www.matecat.com/} and a closely integrated SMT system\footnote{http://www.statmt.org/moses/}
which is already adapted to the task (translation of legal documents).  For
each source sentence, the system proposes an eventual match in the translation
memory and a translation by the SMT system. The human translator can decide to
either post-edit them, or to perform a new translation from scratch.  After one
day of work, we want to use all the post-edited sentences to adapt the SMT
systems, so that the translation quality is improved for the next day. This
means that the SMT system will be adapted to the specific translation project.
One important particularity of the task is that we have a very small amount of
adaptation data, usually around three thousand words per day.

This paper is organized as follows. In the next two sections, we summarize basic
notions of statistical machine translation and continuous space language models.
We then present our tasks and results.
The paper concludes with a discussion and directions of future research.

\section{Related work}

Popular approaches to adapt the LM in an SMT system are mixture models, \eg
\cite{Foster:2007:emnlp_mixadapt,koehn-schroeder:2007:WMT} and data selection.
 In the former case, separate LMs are trained on the available corpora and are
then merged into one, the interpolation coefficients being estimated to
minimize perplexity on an in-domain development corpus. This is known as linear
mixture models. We can also integrate the various corpus-specific LMs as
separate feature functions in the usual log-linear model of an SMT system.

Data selection aims at extracting the most relevant subset of all the available
LM training data. The approach proposed in \cite{Moore:acl10} has turned out
to be the most effective one in many settings. Adaptation of the LM of an SMT
models in an CAT environment was also investigated in several studies, e.g.
\cite{Bach:09,bertoldi-EtAl:2012:WMT,Matecat:2014:mt_journal}.

Adaptation to new data was also investigated in the neural network community,
usually by some type of incremental training on a (subset) of the data.
Curriculum learning \cite{Bengio:09:curriculumlearning}, which aims in
presenting the training data in a particular order to improve generalization,
could be also used to perform adaptation on some new data.
There are a couple of papers which investigate adaptation in the context of a
particular application, namely image processing and speech recognition.  One
could for instance mention a recent work which investigated how to transfer
features in convolutional networks \cite{Yosinski:15:nips_transfer_features},
or research to perform speaker adaptation of a phoneme classifier based on
TRAPS \cite{Trmal:10:nn_spkr_adapt}.

There are also a few publications which investigate adaptation of neural
network language models, most of them very recent.  The insertion of an additional adaption layer to
perform speaker adaptation was proposed by Park et al.
\cite{Park:cslm:intersp10}.  Earlier this idea was explored in \cite{yao2012adaptation} 
for speech recognition through an affine transform of the output layer. Adaptation through data
selection was studied in \cite{jalalvand2013improving} (selection of sentences in out-of-domain corpora based on similarity between sentences) 
and \cite{duh2013adaptation} (training of three models: n-gram, RNN and interpolated LM on two SMT systems: in-domain data only and all-domain).
Several variants of curriculum learning are
explored by Shi et al. to adapt a recurrent LM to a sub-domain, again in the
area of speech recognition \cite{Shia:15:csl_rnn_adapt}. 
Finally, one of the early applications of RNN was in \cite{kombrink2011recurrent}: it was used to rescore the n-best list, speed-up the rescoring process,
adapt an LM and estimate the influence of history.  


\section{Statistical Machine Translation}

In the statistical approach to machine translation, all models are
automatically estimated from examples.  Let us assume that we want to translate
a sentence in the source language $\vs$ to a sentence in the
target language $\vt$.  Then, the fundamental equation of
SMT is, applying Bayes rule:
\begin{equation}
 \vt^* = \arg \max_{\vt} P(\vt|\vs) 
    =  \arg\max_{\vt} P(\vs|\vt) P(\vt)
\end{equation}
The translation model $P(\vs|\vt)$ is estimated from bitexts, bilingual
sentence aligned data, and the language model $P(\vt)$ from monolingual data in
the target language.  A popular approach are phrase-based models which
translate short sequences of words together
\cite{Koehn:2003:hlt_align,Och:2003:ClAlignComp}.  The translation
probabilities of these phrase pairs are usually estimated by simple relative
frequency. The LM is normally a 4-gram back-off model.
The log-linear approach is commonly used to consider more models
\cite{Och:2003:AclMinErr}, instead of just a translation and
language model:
\begin{equation}
  \label{eq:loglinear}
  \vt^* = \arg\max_{\vt} \sum_{m=1}^{M} \lambda_m h_m(\vs, \vt) ,
\end{equation}
\noindent where $h_m(\vs,\vt)$ are so-called feature functions.  The weights
$\lambda_m$ are optimized during the tuning stage.  In the Moses system,
fourteen feature functions are usually used.

Automatic evaluation of an SMT system remains an open question and many
metrics have been proposed. In this study we use the BLEU score which measures
the $n$-gram precision between the translation and a human reference
translation \cite{BLEU:2002:acl}. Higher values mean better translation
quality.

\section{Continuous Space Language Model}
\label{sec:cslm}

\begin{figure}[t!]
  \centering
  \includegraphics[width=0.45\textwidth]{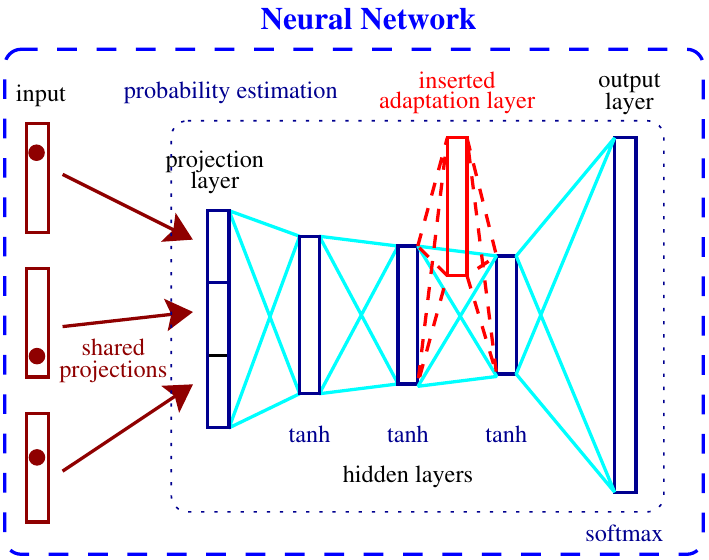}
  \caption{Basic architecture of an CSLM and insertion of an adaptation layer (dashed red).}
  \label{FigCSLM}
\end{figure}

The basic architecture of an CSLM is shown in Figure~\ref{FigCSLM}.
The words are first projected onto a continuous representation, the remaining
part of the network estimates the probabilities.  Usually one tanh hidden and a
softmax output layer are used, but recent studies have shown that deeper
architecture perform better \cite{schwenk:14:nipswshop_nnmt}. We will use three tanh hidden and a softmax output
layer as depicted in Figure~\ref{FigCSLM}.
This type of architecture is now well known and the reader is referred to the
literature for further details, e.g.  \cite{Schwenk:2007:csl}.

All our experiments were performed with the open-source CSLM
toolkit\footnote{The CSLM toolkit is available at
http://www-lium.univ-lemans.fr/\~{ }cslm/} \cite{Schwenk:13:is_cslm}, which
was extended for our purposes. A major challenge for neural network LMs is how
to handle the words at the output layer since a the softmax normalization would
be very costly for large vocabularies. Various solutions have been proposed:
short-lists~\cite{Schwenk:2007:csl}, a class decomposition
\cite{Mikolov:rcslm:asru11} or an hierarchical decomposition
\cite{Le:cslm:icassp11}.  In this work, we use short-lists, but our adaptation
scheme could be equally applied to the other solutions.

\subsection{Adaptation schemes}

As mentioned above, the most popular and most successful adaptation schemes
for standard back-off LMs are data selection and mixture models. Both could be
also applied to CSLMs. In practice, this would mean that we train a completely
new CSLM on data selected by the adaptation process, or that we train several
CSLMs, e.g. a generic and task-specific one, and combine them in linear or
log-linear way.
However, full training of an CSLM usually takes a substantial amount of time,
often several hours or even days in function of the size of the available
training data.  Building several CSLMs and combining them would also increase
the translation time.

Therefore, we propose and compare CSLM adaptation schemes which are very
efficient: they can be performed in a couple of minutes.  The underlying idea
of both techniques is not to train new models, but to slightly change the
existing CSLM in order to account for the new training data.
In the first method, we perform \textbf{continued training} of the CSLM with a
mixture of the new adaptation data and the original training data.
%
In the second method, \textbf{adaptation layers are inserted} in the neural
network as outlined in red in Figure~\ref{FigCSLM}.  This additional layer is
initialized with the identity matrix and only the weights of this layer are
updated. This idea was previously proposed in framework of a speech recognition
system \cite{Park:cslm:intersp10}. We build on this work and explore different
variants of this technique.
An interesting alternative is to keep the original architecture of the NN and
to only modify one layer, e.g. the weights between two tanh layers in
Figure~\ref{FigCSLM}. This variant will be explored in future work.


\section{Task and baselines}

Our task is to improve an SMT system which is closely integrated into an
open-source CAT tool with the post-edits provided by professional human
translators. This tool and algorithms to update standard phrase-based SMT
systems, including back-off language models, were developed in the framework of
the European project MateCat \cite{Matecat:2014:mt_journal}.  We consider the
translation of legal texts from English into German and French.  The available
resources for each language pair are summarized in Table~\ref{TabData}.

\begin{table}[t!]
  \centering
  \begin{tabular}[t]{|c|r|r|}
    \hline
    Corpus & En/German & En/French \\
    \hline
    \hline
    \MC{3}{|l|}{\bf All data:} \\
    Bitexts & 129M & ~512M \\
    Monolingual & 643M & 1300M \\
    \hline
    \MC{3}{|l|}{\bf After data selection:} \\
    Bitexts & ~49M & ~26M \\
    Monolingual & ~44M & 178M \\
    \hline
  \end{tabular}
  \caption{Statistics of the available resources (number of tokenized words)}
  \label{TabData}
\end{table}

\begin{table*}[t!]
\centering
\begin{tabular}{|c|c|c|c|c|}
\hline
\MC{2}{|c|}{LM} & \MC{3}{|c|}{BLEU score} \\
\hline
\hspace{-0mm}Approach\hspace{-0mm} & Adaptation & Dev & Day 1 & Day 2 \\
\hline
\hline
\MC{5}{|c|}{\bf Domain adapted:} \\
\hline
Back-off & n/a & 26.18 & 27.53 & 19.31 \\
CSLM & n/a & 26.89 & 27.14 & 20.28 \\
\hline
\hline
\MC{5}{|c|}{\bf Project adapted;} \\
\hline
Back-off & data selection & 25.76 & (\it 28.45) & 20.14 \\
\hline
& none & 26.45 & (\it 28.65) & 20.57 \\
CSLM & continued training & 26.27 & (\it 33.10) & 21.12 \\
& additional layers & 26.39 & (\it 31.94) & \bf 21.26 \\
\hline
\end{tabular}
\caption{Comparative BLEU scores for the English/German systems.
        Italic values in parenthesis are for information only. They are biased since
        the reference translations are used in training.}
\label{tab:TabResCSLM}
\end{table*}

Each SMT system is based on the Moses toolkit \cite{Moses:2007:acl} and built
according to the following procedure: first we perform data selection on the
parallel and monolingual corpora in order to extract the data which is the most
representative to our development set. In our case, we are interested in the
translation of legal documents.  Data selection is now a well established
method in the SMT community. It is performed for the language and translation
model using the methods described in \cite{Moore:acl10} and
\cite{axelrod-he-gao:2011:EMNLP} respectively.

We train a 4-gram back-off LM and a phrase-based system using the standard
Moses parameters.  The coefficients of the 14 feature functions are optimized
by MERT to maximize the BLEU score on the development data.  This system is
then used to create up to 1000 distinct hypotheses for each source sentence.
We then add a 15th feature function corresponding to the log probability generated by CSLM
 for each hypothesis and the coefficients are again optimized.  This is
usually referred to as \textit{n-best list rescoring}.
We call this final system \textbf{domain-adapted} since it is optimized to
translate legal documents.  This system is then used to assist human
translators to translate a large document in the legal domain.

\vspace{\baselineskip}
Typically, we will process day by day: after one day of work, all the human
translations (created from scratch or by post-editing the hypotheses from the
SMT system) are injected into the system and we hope that SMT will perform
better on the rest of the document to be translated, e.g.  on the second day of
work. This procedure can be repeated over several days when the document is
rather large (see section~\ref{SectEnFr}).  Usually humans are able to
translate approximately 3~000 words per day.  We call this procedure
\textbf{project-adaptation}.


\begin{table*}[t!]
  \centering
  \begin{tabular}{|c|c|c|c|c|}
    \hline
    Percentage of & Generic data & Day~1 data & \# examples & training time \\
    adaptation data & (44M words) & (3.2k words) & per epoch & per epoch \\
    \hline
    \MC{5}{|l|}{\textbf{Domain-adapted CSLM:}} \\
\hline
      none	& 19.3M (42\%)  & n/a & 19.3M & 3250 sec \\ 
    \hline
    \MC{5}{|l|}{\textbf{Project-adapted CSLM:}} \\
\hline

      14\% & 19~356 (0.042\%) & 3~220 & 22~576 & 3.5 sec \\	
      25\% & ~9~696 (0.021\%) & 3~220 & 12~916 & 2.0 sec \\	
      45\% & ~3~899 (0.008\%) & 3~220 & ~7~119 & 1.1 sec \\	
      62\% & ~1~967 (0.004\%) & 3~220 & ~5~187 & 0.6 sec \\ 
      77\% & ~1~003 (0.002\%) & 3~220 & ~4~223 & 0.5 sec \\	
    \hline
  \end{tabular}
  \caption{English/German system: number of examples (28-grams) seen by the CSLM at each epoch.
	For the domain adapted system, we randomly resample about 42\% of the examples at each epoch.
	For the project-adapted system, we experimented with various mixtures between generic and project
	specific data (Day~1). We don't want to train on Day~1 data only since this would result in
	strong over-fitting.
  }
  \label{TabResample}
\end{table*}

\subsection{Results for the English/German system}

The 4-gram back-off LM built on the selected data has a perplexity of 151.1 on
the domain-specific development data.  Given the fact that an CSLM can be very
efficiently trained on long context windows, we used a 28-gram in all
experiments. By these means we hope to capture the long range dependencies of
German.
%
The projection layer of the CSLM was of dimension 320, followed by three tanh
hidden layers of size 1024 and a softmax output layer of 32k neurons (short-list).
This short-list accounts for around 92 \% of the tokens used in the corpus.   
The initial learning rate was set to 0.06 and exponentially decreased over the
iterations.  The network converged after 7 epochs with a perplexity of 96.6,
\ie a 36\% relative reduction.  The total training time is less than 7~hours on
a Nvidia K20x GPU.
Table~\ref{tab:TabResCSLM} (upper part) gives the BLEU score of these baseline
domain-adapted systems.

To analyze our project adaptation techniques we have split another legal
document into two part, \textit{``Day~1''} and \textit{``Day~2''}. The first part, \textit{``Day~1''}, containing around 3.2K words, is used to
adapt the SMT system and the CSLM, aiming to improve the translation
performance on the second part, named \textit{``Day~2''}. Note that the
performance on \textit{``Day~1''} itself, after adaptation, is of limited
interest since we could quite easily overtrain the model on this data.  On the
other hand, it is informative to monitor the performance on the domain-generic
development set. Ideally, we will improve the performance on
\textit{``Day~2''}, \ie future text of the same project than the adaptation
data, with only a slightly loss on the generic development data.

Various adaptation schemes are compared in Table~\ref{tab:table1}.
The network is adapted on the data from \textit{Day~1} and we
want to improve performance on \textit{Day~2}.  At the same time, we do not
want to overfit the data and keep good performance on the domain-specific Dev
set. To achieve this, we continued training of the networks with a mixture of
old and new data.
All the adaptation data was always used (\textit{Day~1}, 3.2k words) and small
fractions of the domain-selected data were randomly sampled at each epoch, so
that the adaptation data accounts for 14, 25, 45, 62 and 77 \% respectively.
Since the networks are trained on very small amounts of data (4 - 23k words),
the overall adaptation process takes only a few minutes.  The statistics of the
data used at each epoch is detailed in Table~\ref{TabResample}.  We will show
below that it is important to perform the adaptation of the CSLM with a mixture of
generic and adaptation data to prevent overfitting.

\begin{table*}[t!]
\begin{center}
\begin{tabular}{|c|c|c|c|c|c|c|}
\hline
Network & Updated & Activation & Addtl. & Percentage of & \MC{2}{c|}{Perplexity} \\
architecture& layers & function & params & adapt. data & Day 2 & Dev \\
\hline
\hline
\MC{7}{|l|}{\bf Original network architecture:} \\[4pt]
\hline
\begin{tabular}[c]{c}
1024-1024-1024 \\ without adaptation
\end{tabular}
&-&Tanh&-&-&126.1&96.6\\
\hline
\hline
\MR{3}{*}{1024-1024-1024 }&\MR{5}{*}{All}&\MR{5}{*}{Tanh}&\MR{5}{*}{-}&14\%&$\textbf{94.6}^*$&98.7\\
\MR{3}{*}{with incremental training}&&&&25\%&103.7&~97.3\\[4pt]
&&&&45\%&102.9&~98.9\\
&&&&62\%&102.7&100.2\\
\hline
\hline
\MC{7}{|l|}{\bf Insertion of an adaptation layer:} \\
\hline
\MR{2}{*}{1024-$\textbf{1024}$-1024-1024}&inserted&\MR{2}{*}{Linear}&\MR{2}{*}{1M}&14\%&106.0&~97.4\\
&one only&&&25\%&104.9&~99.5\\
\hline
\MR{2}{*}{1024-1024-$\textbf{1024}$-1024}&inserted&\MR{2}{*}{Linear}&\MR{2}{*}{1M}&14\%&103.8&98.8\\
&one only&&&25\%&97.9&102.5\\
\hline
\MR{2}{*}{1024-1024-1024-$\textbf{1024}$}&inserted&\MR{2}{*}{Linear}&\MR{2}{*}{1M}&14\%&101.2&100.8 \\
&one only&&&25\%&102.2&104.1\\
\hline
\hline
\MR{2}{*}{1024-$\textbf{1024}$-1024-1024}&inserted&\MR{2}{*}{Tanh}&\MR{2}{*}{1M}&14\%&105.7&~96.8\\
&one only&&&25\%&104.6&~98.9 \\
\hline
\MR{2}{*}{1024-1024-$\textbf{1024}$-1024}&inserted&\MR{2}{*}{Tanh}&\MR{2}{*}{1M}&14\%&103.5&~96.4\\
&one only&&&25\%&102.6&~98.4\\
\hline
\MR{2}{*}{1024-1024-1024-$\textbf{1024}$}&inserted&\MR{2}{*}{Tanh}&\MR{2}{*}{1M}&14\%&101.5&$\textbf{95.1}^*$\\
&one only&&&25\%&101.3&97.4 \\
\hline

\end{tabular}
\end{center}
\caption{Perplexities of CSLMs with one new hidden layer adapted to \textit{Day~1}. Bold values in the architecture column are the new hidden layers. Bold values in the last two columns are the best perplexities for the respective test corpora. Tanh is a shorthand notation for the hyperbolic tangent activation function. Percentage is the proportion of \textit{Day~1} data in the total corpora (see Table~\ref{TabResample}). 
  All networks have been trained for 50 iterations.}
\label{tab:table1}
\end{table*} 

\vspace{\baselineskip}
We experiment along the following lines:
\begin{enumerate}
  \item different resampling coefficients of adaptation and generic data according to Table~\ref{TabResample}.
  \item network topologies:
  \begin{itemize}
     \item[a)] continue training of the original network updating all the weights.
     \item[b)] insert one or two hidden layers with 1024 neurons using linear
or hyperbolic tangent activation functions respectively. These additional layers are
initialized with the identity matrix and only these layers are updated using backpropagation function.
  \end{itemize}
  %
\end{enumerate}

We record the perplexity of the adapted CSLM on \textit{Day~2} ($\sim 11K$
words), which is then used as a guideline for selecting the best networks to
integrate into an SMT system (marked with an asterisk in the Table \ref{tab:table1}).
Lowest perplexity was obtained by keeping the baseline network topology (upper part of Table~\ref{tab:table1}) 
when \textit{Day~1} data constituted 14 \% of the
incremental training data set: the perplexity on \textit{Day~2} decreases from 126.1 to
94.6, with a minor increase on the Dev set (96.6$\ra$98.7). Using larger
fractions of \textit{Day~1} leads to over-fitting of the network: the perplexity
on \textit{Day~2} and the generic Dev set increases.

The lower part of Table~\ref{tab:table1} summarizes the results when inserting
one \textit{adaptation layer}, with a linear or tanh activation function, at
three different slots respectively. For each configuration, we explored five
different proportions of the baseline corpora and \textit{Day~1} (cf.
Table~\ref{TabResample}), but for clarity, we only report the most interesting
results. The overall tendency was that using more than 25\% of
\textit{Day~1} systematically leads to over-fitting of the network.
Several conclusions can be made:
a) an tanh adaptation layer outperforms a linear one;
b) it is better to insert the adaptation layer at the end of the network;
c) updating the weights of the inserted layer only overfits less than incremental
training the whole network (comparing the last block in Table~\ref{tab:table1} with
the second block): the perplexity on \textit{Day~1} decreases substantially (126.1$\ra$101.5)
and we observe a slight improvement on the Dev set (96.6$\ra$95.1).

Finally, Table \ref{tab:TabResCSLM} lower part gives the BLEU scores of the
project-adapted systems. When no CSLM is used, the BLEU score on Day~2
increases from 19.31 to 20.14 (+0.83). This is achieved by adapting the
translation and back-off LM (details of the algorithms can be found in
\cite{Matecat:2014:mt_journal}). Both CSLM adaptation schemes obtained quite
similar BLEU scores: 21.12 and 21.26 respectively, the insertion of one
additional tanh layer having a slight advantage.  Overall, the adapted CSLM
yields an improvement of 1.12 BLEU (20.14 to 21.26) while it was about 1 point
BLEU for the domain-adapted system (19.31 to 20.28).  This nicely shows the
effectiveness of our adaptation scheme, which can be applied in a couple of
minutes.


\subsection{Results for the English/French system}
\label{SectEnFr}

A second set of experiments was performed to confirm the effectiveness of our
adaptation procedure on a different language pair: English/French. In the MT
community it is well known that the translation into German is a very hard
task which is reflected in the low BLEU scores around 20 (see
Table~\ref{tab:TabResCSLM}).  On the other hand, our baseline SMT system for
the English/French language pair has a BLEU score well above 40. One may argue
that it is more complicated to further improve such a system.


\begin{table*}[t!]
\centering
\begin{tabular}{|l|c|c|c|c|c|}
\hline
Approach & Day 1 & Day 2 & Day 3 & Day 4 & Day 5 \\
\hline
\hline
\MC{6}{|l|}{\bf Baseline SMT system:} \\ 
back-off LM& 48.84/63.69 & 44.07/62.13 & 46.88/67.14 & 43.22/64.74 & 47.77/67.07\\
      CSLM & 52.25/67.04 & 46.61/65.64 & 49.73/70.70 & 45.68/68.61 & 50.06/69.70 \\
\hline
\MC{6}{|l|}{\bf Adapted SMT system:} \\ 
baseline CSLM & \MR{4}{*}{n/a} & 52.01/66.68 & 57.35/75.31& 54.99/71.88&59.11/74.49\\
adapted CSLM  & &54.61/67.97&60.23/75.90&57.19/72.05&61.83/5.21\\
\cdashline{3-6}
Improvement obtained&&\MR{2}{*}{2.60/1.29}&\MR{2}{*}{2.88/0.56}&\MR{2}{*}{2.20/0.17}&\MR{2}{*}{2.72/0.72}\\
by adapted CSLM&&&&&\\
\hline   
\end{tabular}
\caption{BLEU scores obtained by a baseline SMT (without and with an CSLM) and a project-adapted SMT with baseline (unadapted) CSLM and adapted CSLM.
 The first value in every cell is the BLEU score obtained with respect to the reference translation of the human translator;
 the second one is calculated with respect to all the 3 references created by the professional translators (\ie~obtained by post-edition) and an independent reference.}
\label{tab:table6}
\end{table*}

In addition, we investigate adaptation of the SMT system and the CSLM over five
consecutive days: the human translator works for one day and corrects the SMT
hypothesis, these corrections are used to adapt the system for the second day. 
Human corrections are again inserted into the system
and a new system for the third day is built, and so on.  With this adaptation
scheme we want to verify whether our methods are robust or quickly overfit the
adaptation data.
The number of words for each day are about three thousand.
%
A 16-gram CSLM for the French target language with a short-list of 12k was used. Training was performed for 15 epochs.

\begin{table}[h!]
\centering
\begin{tabular}{|c||c|c|c|c|}
\hline
Day & Day 1 & Days 1-2& Days 1-3 & Days 1-4\\
\hline
1&39 \%&27.9 \%&21.6 \%& 17.7 \% \\
2&-&29.6 \%&22.9 \%&18.8 \% \\
3&-&-& 22.3 \%& 18.1 \% \\
4&-&-&-& 17.4 \%\\
\hline
\end{tabular}
\caption{English/French task: proportion of each day in the adaptation data set,
\eg at the end of Day~2, we create an adaptation corpus which consists of 27.9\% 
and 29.6\% of data from Day~1 and Day~2 respectively,
the remaining portions are randomly resampled in the training data.}
\label{tab:tab_proportion}
\end{table}

For this task, we only used the incremental learning method (see Table~\ref{tab:table1})
as it yielded the lowest perplexity in the English/German experiment.
%
The data from the five consecutive days is coming from one large document which
is assumed to be from one domain only.  Therefore, we decided to always use all
the available data from the preceding days to adapt our models.  For instance,
after the third day, the data from Day~1, 2 and 3 is used to build a new system
for the fourth day.
The proportions of each day in the corpus used to continue the training of the
CSLM are given in Table~\ref{tab:tab_proportion} (note that every day's
proportion decreases, but their combined share increases from $39 \%$ to $68
\%$).
The perplexities of the various CSLMs are given in Table~\ref{tab:table5}.

\begin{table}[h!]
\centering
\begin{tabular}{|c||c|c|}
\hline
Data & CSLM baseline & CSLM adapted \\
\hline
Day 1 & 233.9 & - \\
Day 2 & 175.6 & 130.3 \\
Day 3 & 153.0 & 130.2 \\
Day 4 & 189.4 & 169.4 \\
Day 5 & 189.2 & 167.7 \\
\hline
\end{tabular}
\caption{English/French task: perplexities of baseline and adapted CSLM (on all preceding days), \eg the CSLM tested on Day~4 is the baseline CSLM that had been adapted with Days~1-3.}
\label{tab:table5} 
\end{table}       

One first observation is the rather high perplexity of the models on each day.
This shows the importance of project adaptation even when domain related data is available.
Adaptation allows to decrease the perplexity by more than 10\% relative for each day.
While the perplexities vary between the project days, they are reduced in every case, 
which demonstrates the effectiveness of the adaptation method.

In order to evaluate the impact of the CSLM adaptation on the SMT system, 
we performed various translation experiments.
The results are provided in Table~\ref{tab:table6}. The BLEU scores of the various
systems using the baseline and the adapted CSLMs are presented.  
We run tests with three different human translators - for the sake of clarity, we
provide detailed results for one translator only.
The observed tendencies are similar for the two other translators.
First of all, one can see that the CSLM improves the BLEU score of the baseline
systems between 2.3 to 3.4 BLEU points, \eg for Day~2 from 44.07 to 46.61.
Adapting the whole SMT system to the new data improves significantly the
translation quality, \eg from 46.61 to 52.01 for Day~2, without changing the
CSLM.  The proposed adaptation scheme of the CSLM achieves additional important
improvements, in average 2.6 BLEU points. This gain is relatively constant for
all days.

For comparison, we also give the BLEU scores when using four reference
translations: the one of the three human translators and one independent
translation which was provided by the European Commission.

We still observe some small gains although three out of four translations were not used
in the adaptation process. This shows that our adaptation scheme not only learns
the particular style of one translator, but also achieves more \texttt{generic} improvements.
This also shows that the adaptation process is beneficial for improving state-of-the-art systems 
which already perform very well on certain tasks.


\section{Conclusions}

In this paper, we presented a thorough study of different techniques to adapt a
continuous space language model to small amounts of new data. In our case, we
want to integrate user corrections so that a statistical machine translation
system performs better on similar texts.
Our task, which corresponds to concrete needs in real-world applications, is
the translation of a document by an human over several days. The human
translator is assisted by an SMT system which proposes translation hypothesis
to speed up his work (post editing). After one day of work, we adapt the CSLM
to the translations already performed by the human translator, and show that
the SMT system performs better on the remaining part of the document.

We explored two adaptation strategies: continued training of an existing neural
network LM, and insertion of an adaptation layer with the weight updates being
limited to that layer only.  In both cases, the network is trained on a
combination of adaptation data (3--15k words) and a portion of similar size,
randomly sampled in the original training data.  By these means, we avoid
over-fitting of the neural network to the adaptation data.  Overall, the
adaptation data is very small -- less than 50k words -- which leads to very
fast training of the neural network language model: a couple of minutes on a
standard GPU.

We provided experimental evidence of the effectiveness of our approach on two
large SMT tasks: the translation of legal documents from English into German
and French respectively.  In both cases, significantly improvement of the
translation quality was observed.

\bibliographystyle{acl}
\bibliography{strings_short,mt_hs,speech_hs}
\end{document}